\documentclass{article}

     \PassOptionsToPackage{numbers, compress}{natbib}

 \usepackage[preprint]{neurips_2020}

\usepackage{natbib}
\usepackage[utf8]{inputenc} 
\usepackage[T1]{fontenc}    
\usepackage{url}            
\usepackage{booktabs}       
\usepackage{amsfonts}       
\usepackage{nicefrac}       
\usepackage{microtype}      
\usepackage{url}
\usepackage{graphicx} 
\usepackage{caption}
\usepackage{mwe}
\usepackage{color}
\usepackage{array}
\usepackage{wrapfig}
\usepackage{amsmath}
\usepackage{amssymb,amsthm}
\usepackage{graphicx}
\usepackage{algorithm}
\usepackage{algorithmic}
\usepackage{graphicx}
\usepackage{subfig}
\usepackage{graphics}
\usepackage{epsfig}
\usepackage{epstopdf}
\def\muu{\boldsymbol\mu}
\def\p{\mathbf p}

\def\W{\mathbf{W}}

\def\R{\mathbb{R}}

\def\A{\mathbf{A}}

\def\G{\mathcal{G}}
\def\X{\mathbf{X}}
\def\T{\mathbf{T}}
\def\p{\mathbf{p}}
\def\U{\mathbf{U}}

\def\b{\mathbf{b}}

\def\M{\mathbf{M}}
\def\z{\mathbf{z}}
\def\H{\mathbf{H}}

\def\Z{\mathbf{Z}}
\def\P{\mathbf{P}}
\newtheorem{theorem}{Theorem}
\usepackage{graphicx}
\usepackage{caption}

\usepackage{mathtools}

\title{Structural Landmarking and Interaction Modelling: on Resolution Dilemmas in Graph Classification}

\author{%
  Kai Zhang\\
  Dept. Computer \& Information Sciences\\
  Temple University\\
  Philadelphia, PA 19122 \\
  \texttt{zhang.kai@temple.edu} \\
   \And
   Yaokang Zhu, Jun Wang \\
   School of Computer Science and Technology\\
   East China Normal University\\
   Shanghai, China \\
   \texttt{
52184501026,
jwang@sei.ecnu.edu.cn} \\
   \AND 
   Jie Zhang\\
   Institute of Brain-Inspired Intelligence\\
   Fudan University\\
   Shanghai, China \\
   \texttt{jzhang080@gmail.com} \\
  \And
   Hongyuan Zha\\
   College of Computing\\
   Georgia Institute of Technology\\
   Atalanta, GA 30332 \\
  \texttt{zha@cc.gatech.edu} \\
}

\begin{document}

\maketitle

\begin{abstract}
Graph neural networks are promising architecture for learning and inference with graph-structured data. Yet difficulties in modelling the ``parts'' and their ``interactions'' still persist in terms of graph classification, where  graph-level representations are usually obtained by squeezing the whole graph into a single vector through graph pooling. From complex systems point of view, mixing all the parts of a system together can affect both model interpretability and predictive performance, because properties of a complex system arise largely from the interaction among its components.
We analyze the intrinsic difficulty in graph classification under the unified concept of ``resolution dilemmas'' with learning theoretic recovery guarantees, and propose ``\emph{SLIM}'', an inductive neural network model for Structural Landmarking and Interaction Modelling. It turns out, that by solving the resolution dilemmas, and  leveraging explicit interacting relation between component parts of a graph to explain its complexity, 
 \emph{SLIM} is more interpretable, accurate, and offers new insight in graph representation learning.
\end{abstract}

\section{Introduction}

 



Complex systems are ubiquitous phenomenon in natural and scientific disciplines, and how relationships between parts give rise to global behaviours of a system is a central theme in many areas of study such as system  biology \cite{biology}, neural science \cite{brain}, and drug and material discoveries \cite{drug}  \cite{material}.

Graph neural networks are promising architecture for representation learning on graphs - the structural abstraction of  complex system. 
State-of-the-art performance is observed in various graph mining tasks \cite{GCN2,GCN5,graphSAGE,GNNpower,gat,WLneural,GNNreview, GNNreview2,GNNreview3}.
However, due to the non-Euclidian nature, challenges still exist in  graph classification. 
For example, to generate a fixed-dimensional graph-level representation, GNN combines information from each node through \emph{graph pooling}. In combined forms, a graph will collapse into a ``super-node'', where identities of the constituent sub-graphs and their inter-connections are mixed together. Is this the best way to generate graph-level features? From complex systems view, mixing all parts of a system  can affect interpretability and model prediction, because properties of a complex system  arise largely from the \emph{interactions} among its components \citep{molecular,book_complex,book_complex2}.


The choice of the ``collapsing''-style  graph pooling roots deeply in the lack of natural alignment among graphs that are not isomorphic. Therefore the pooling sacrifices structural details for feature compatibility. In recent years, substructure patterns draw considerable attention in graph mining, such as motifs  \citep{motif1,motif2,motif3,motif4} and graphlets \cite{fast-gkernel}. It provides an intermediate scale for  structure comparison or counting, and has been considered in node embedding \cite{motif_embed}, deep graph kernels \cite{Deep-gkernel} and graph convolution \citep{GNNmotif1}.
However, due to the combinitorial nature, only substructures of very small sizes (4 or 5 nodes) can be considered \cite{Deep-gkernel, motif3}, greatly limiting the coverage of structural variations; also, handling  substructures as discrete objects makes it difficult to compensate for their similarities, at least computationally, and so the risk of overfit may rise in supervised learning scenarios. 

These intrinsic difficulties are related to the concept of \emph{resolution} in graph-structured data processing. Resolution is the scale at which measurements can be made and/or information processing algorithms are conducted. 
Here, we will first define two relevant terms, i.e., the spatial resolution and the structural resolution, and how they may affect the performance of graph classification. 

First, \emph{ {spatial resolution}} is related to the  geometrical scale of the ``elementary component'' of a graph on which an algorithm operates. It can range from nodes, to sub-graphs, or entire graph. Graph details beyond effective spatial resolution are algorithmically unidentifiable. For example, graph pooling compresses the whole graph into a single vector, and so the spatial resolution drops to the lowest: node and edge identities are mixed together, and subsequent classification layer can no longer exploit any substructure or their connections, but just a global aggregation.  We call this \textbf{vanishing spatial resolution}.  
Insufficient spatial resolution may affect the interpretability, and also the predictive power since global property of a complex system arises largely from the its inherent interactions \citep{molecular,book_complex,book_complex2}.



Second, \emph{{structural resolution}} is the fineness level in differentiating between substructures. substructures (or sub-graphs) shed light on functional organization and graph alignment. However, they are treated in a discrete, and over-delicate manner: in exact matching, two substructures are considered distinct even if they share significant similarity. We call it \textbf{exploding structural resolution}. It can lead to risk of overfitting, similar to observed in deep graph kernels \citep{Deep-gkernel} and dictionary learning \cite{adpt_size}. 

We believe that both  {resolution dilemmas} 
originate from the way we perform profiling, identification, and alignment of substructures. Substructures are building blocks of a graph; relations like interaction or alignment are all defined between substructures (of varying scales). However, exact substructure matching is too costly and prone to overfit, leading to exploding structural resolution; meanwhile, graph alignment becomes infeasible when substructure matching is poorly defined, and so collapsing-style graph pooling becomes the norm, which finally leads to vanishing spatial resolution.

\textbf{Our contribution}. In this paper, we propose a simple neural architecture called  ``{S}tructural {L}andmarking and {I}nteraction {M}odelling'' - or SLIM, for inductive graph classification.
The key idea is to embed substructure instances into a continuous metric space and learn structural landmarks there for explicit interaction modelling. The SLIM network can effectively resolve the resolution dilemmas. More importantly, by fully exploring the diverse structural distribution of the input graphs, any substructure instance and even unseen examples can be mapped parametrically to a common and optimizable structural landmark set. This enables a novel, \emph{identity-preserving graph pooling} paradigm, where the interacting relation between constituent parts of a graph can be modelled explicitly, shedding important light on the functional organizations of complex systems.



The design philosophy of SLIM comes from the long-standing views of complex systems: complexity arises from interaction.  Therefore, explicit modelling of the parts and their interactions is key to explaining the complexity and improving the prediction.  In contrast, graph neural networks is more about ``integration'', where delicate part-modelling like convolution does exist but finally obscured in the pooling process. 
It turns out, that by respecting the structural organization of complex systems, SLIM is more interpretable, accurate, 
and provides new insights in graph representation learning. 


We will discuss the resolution dilemmas and related works in Section~\ref{sec:2}. Section~\ref{sec:3}, ~\ref{sec:theory} and ~\ref{sec:exp} covers the design, analysis, and performance of SLIM, respectively. The last section concludes the paper.

\section{Resolution Dilemmas in Graph Classification}
\label{sec:2}
A complex system is composed of many parts that interact with each other in a non-simple way. 
Since graphs are structural abstraction of complex systems, accurate graph classification depends on how global properties of a system relate to its structure. 
It is believed that the property (and complexity) of a complex system arises from the interaction among its components \cite{book_complex,book_complex2}. So, accurate interaction modelling should benefit prediction. However, this is non-trivial  due to resolution dilemmas. 

\subsection{Spatial Resolution Diminishes in Graph Pooling}
Graph neural networks (GNN) for graph classification  typically has two stages: graph convolution and graph pooling \citep{graphSAGE,GNNpower}. The spatial resolutions for these two stages are significantly different. 

The goal of convolution is to pass message among neighboring nodes in the general form of $h_v = \texttt{AGGREGATE}\left(\{h_u, u\in \mathcal{N}_v\}\right)$, where $\mathcal{N}_v$ is the neighbors of $v$ \cite{graphSAGE,GNNpower}.
Here, the spatial resolution is controlled by the number of convolution layers: more layers capture lager substructures/sub-trees   and can lead to improved discriminative power \cite{GNNpower}. In other words, a medium resolution (substructure level) can be more informative functional markers than a high resolution (node level). In practice, multiple resolutions can be combined via \texttt{CONCATENATE} function \cite{graphSAGE,GNNpower} for subsequent processing.

The goal of graph pooling is to generate compact,  graph-level representations that are compatible across graphs. Due to the lack of natural alignment between graphs that are not isomorphic, graph pooling typically ``squeezes'' a graph $\G$ into a single vector (or ``super-node'') in the form of $h_\G = \texttt{READOUT}\left(\{f(h_v), \forall v\in \mathcal{V}\}\right)$, where $\mathcal{V}$ is the node of $\G$. Different readout functions have been proposed, including max-pooling \citep{max_pooling}, sum-pooling \cite{GNNpower}, various pooling functions (\texttt{MEAN}, \texttt{LSTM}, etc.) \cite{graphSAGE}, or deep sets \citep{deep_set}; attention has been used to evaluate node importance in attention pooling \citep{att_pool} and gPool \citep{unet};  besides, hierarchical differential pooling has also been investigated   \citep{dif_pool}. 

\begin{figure}[htb]
\begin{center}
    \includegraphics[height=1.8in,width=5.3in]{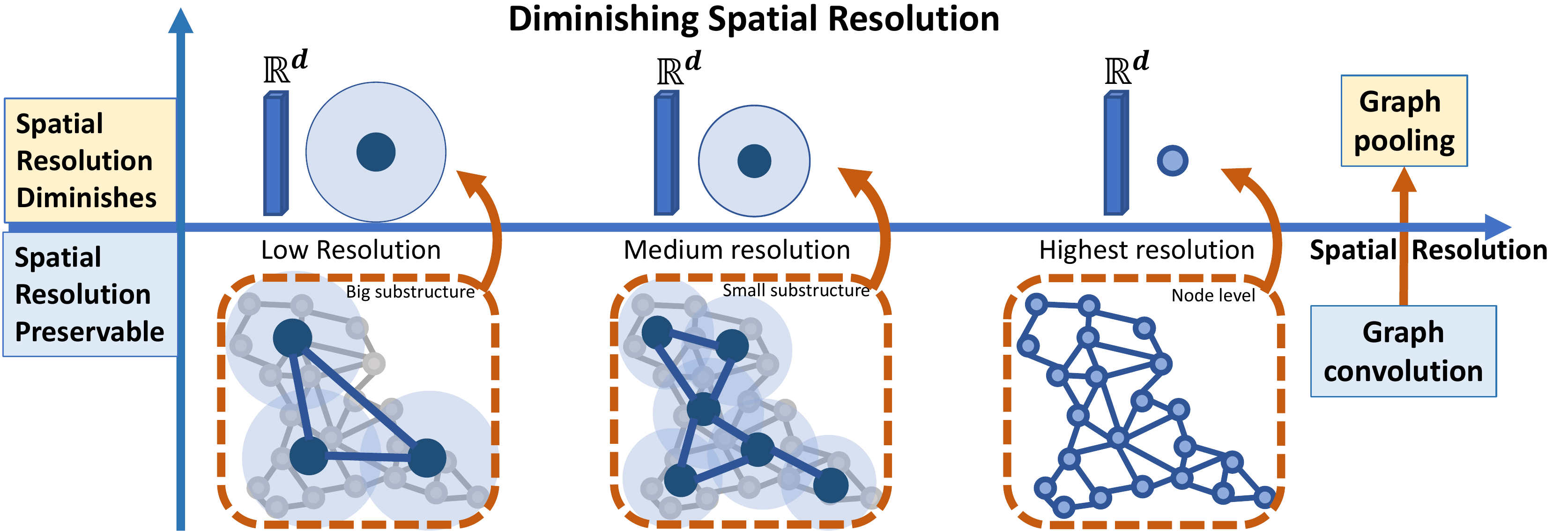}
\caption{Spatial resolution vanishes after graph pooling. \small{(Note: not all nodes are marked with convolution - the shaded circles; see Appendix Sec~8.4 for more discussion on relation with hierarchical processing.)}}\label{fig:spa_res}
\end{center}
\end{figure}

An important resolution bottleneck occurs in graph pooling, as shown in Figure~\ref{fig:spa_res}.
Since all the nodes are mixed into one, subsequent classifier can no longer identify any individual substructure nor their interactions, regardless of the resolution in graph convolution. 
We call this ``diminishing spatial resolution'', which can be undesirable\footnote{Some work adopt different aggregation strategies: Sortpooling arranges nodes in a linear chain and perform 1d-convolution \cite{DGCNN}; SEED uses distribution of multiple random walks \cite{SEED}; Deep graph kernel evaluates graph similarity by subgraph counts \cite{Deep-gkernel}. Explicit modelling of the interaction between graph parts is not considered.} in that: (1) how much information in well-designed convolution domain can penetrate through the pooling layer for final prediction is hard to analyze/control; (2) in  molecule  classification, graph labels hinge on functional modules and how they organize \cite{drug}; an overly coarse spatial resolution will mix up functional modules and conceal their interaction.

Can meaningful spatial resolution(s) survive graph pooling? The answer is yes. Indeed, it involves substructure alignment, and the notion of structural resolution. See discussions below. 






\subsection{Structural Resolution Explodes in Substructure Identification}

Substructures are the basic unit to accommodate interacting relations. A global criteria to identify and align substructures is the key to preserving substructure identities and comparing the inherent interactions across graphs. Again,  the fineness level in determining whether two substructures are ``similar'' or ``different'' is subject to a wide spectrum of choices, which we call ``structural resolution''.

\begin{figure}[htb]
\begin{center}
    \includegraphics[height=1.75in,width=5.15in]{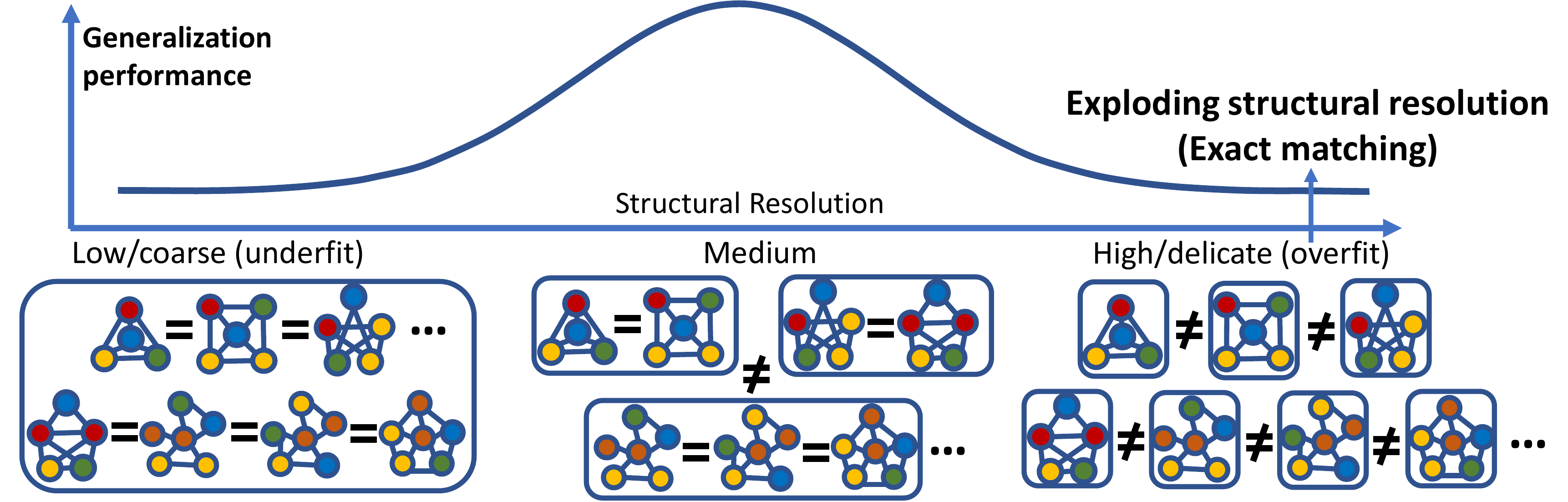}
\caption{How structural resolution may affect the generalization performance. Only small  substructures here for illustration; node types do  make a difference in profiling the substructures. }\label{fig:str_res}\end{center}
\end{figure}

We illustrate in Figure~\ref{fig:str_res}. The right end denotes the finest resolution in differentiating between substructures: exact matching, as we  manipulate motif/graphlet \cite{motif1,motif2,motif3,GNNmotif1,fast-gkernel}. The exponential configuration of sub-graphs will finally lead to an ``exploding'' structural resolution, because maintaining a large number of unique substructures is infeasible and easily overfits. The left end of the spectrum treats all substructures the same and underfits the data. We are interested in a medium structural resolution, where similar substructures are mapped to the same identity, which we believe can benefit the generalization performance (see Figure~\ref{fig:dummy} for empirical evidence).


Theoretically, an over-delicate structural resolution corresponds to a highly ``coherent'' basis in representing a graph, leading to unidentifiable dictionary learning \cite{ERC,supervised_dic}. 
Structural landmarking is exactly aimed at controlling structural resolution and improve  incoherence for graph classification.


\section{Structural Landmarking and Interaction Modelling (SLIM)}
\label{sec:3}
 
Considering the difficulty in manipulating substructures as discrete objects, we
embed them in a continuous space, and transform all structure-related operations from discrete and off-the-shelf version to continuous and optimizable counterpart. The key idea of  SLIM  is the identification of structural landmarks in this new space, via both unsupervised compression and supervised fine-tuning, through the distribution of embedded  substructures under possibly multiple scales. Structural landmarking resolves resolution dilemmas and allow explicit interaction modelling in graph classification. 

\textbf{Problem Setting}. Give a set of labeled graphs $\{\G_i, y_i$\}'s for $i = 1,2,..., n$, with each graph defined on the node/edge set $\G_i = (\mathbf{V}_i,\mathbf{E}_i)$ with adjacency matrix $\A_i\in\R^{n_i\times n_i}$ where $n_i = |\mathbf{V}_i|$, and  $y_i\in\{\pm 1\}$. 
Assume that nodes are drawn from $c$ categories, and the node attribute matrix for $\G_i$ is $\X_i\in\R^{n_i\times c}$. Our goal is to train an inductive model to predict the labels of the testing graphs.  

\begin{figure}[htb]
\begin{center}\vskip -2mm
\includegraphics[width=1\textwidth]{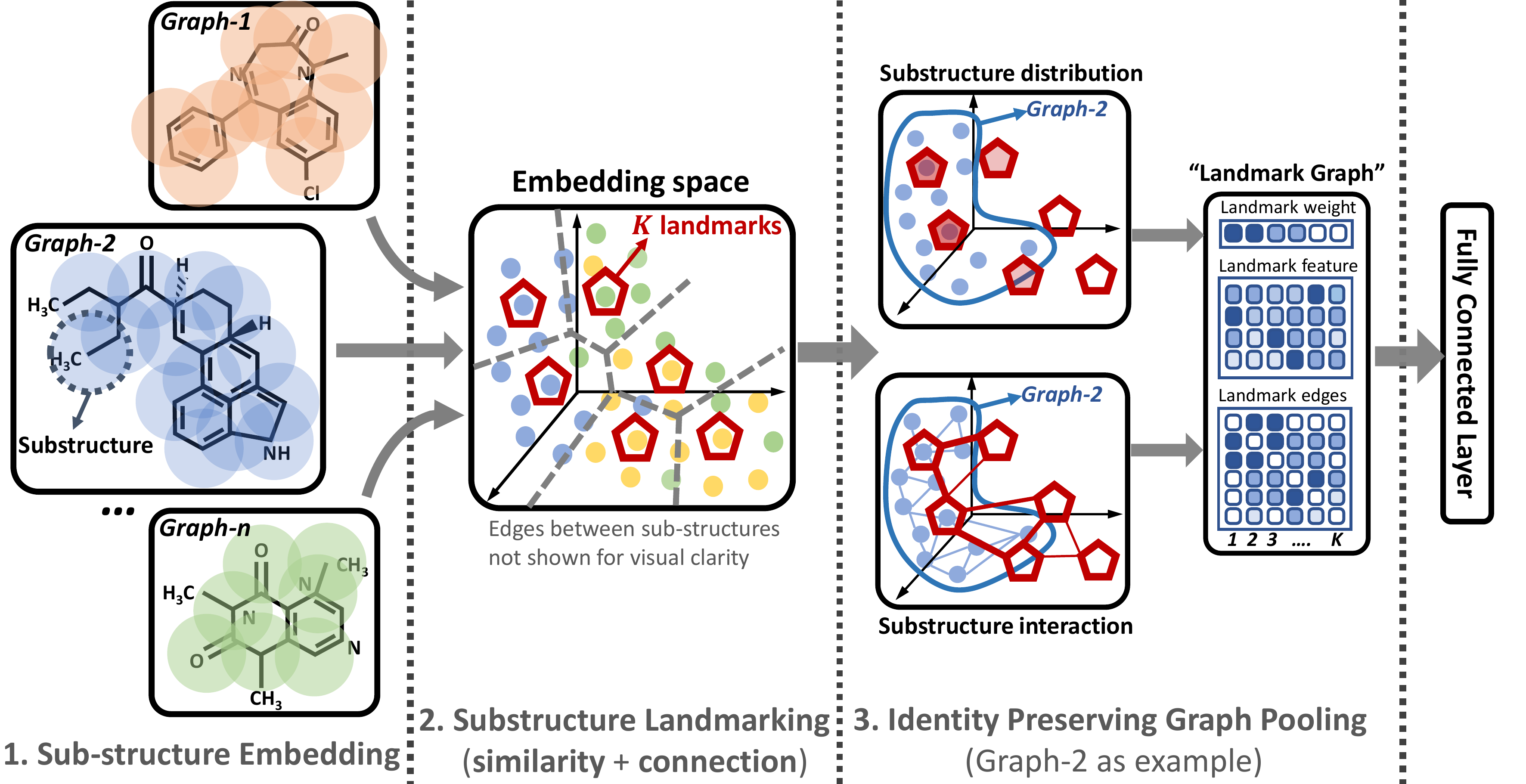}\caption{The three main steps of the SLIM network illustrated in molecule graph classification. }\label{fig:slim}
\end{center}
\end{figure}
The SLIM network has three main steps: (1) sub-sturcture embedding, (2) substructure landmarking, and (3) identity-preserving graph pooling, as shown in Figure~\ref{fig:slim}. Detailed discussion follows.

\subsection{Substructure Embedding}
The goal of substructure embedding is to extract substructure instances and embed them in a metric space. One can employ multiple layers of convolutions \cite{graphSAGE,GNNpower} to model substructures (rooted sub-trees), or randomly sample sub-graphs \cite{fast-gkernel}. For convenience, 
we simply extract one sub-graph instance from each node using a $k$-hop breath-first search, which controls  the spatial resolution\footnote{When $k$ is large, one subgraph around each node may be unnecessary. See discussion in Appendix (Sec8.4). }. 
In Figure~\ref{fig:slim}, sub-graphs  in the shaded circles around each atom is a substructure instance.

Let $\A_i^{(k)}$ be the $k$th-order adjacency matrix, i.e., the $pq$th entry equals 1 only if node $p$ and $q$ are within $k$-hops away. Since each sub-graph is associated with one node, the sub-graphs extracted from $\G_i$ can be represented as $\Z_i = \A^{(k)}_i\X_i$, whose $j$th row is a $c$-dimensional vector summarizing the counts of the $c$ node-types in the sub-graph around the $j$th node. Variations include (1) emphasize the center node, $\Z_i = [\X_i;\; \A_i\X_i]$; (2) layer-wise node distribution $\Z_i = [\tilde\A^{(1)}_i\X_i; \; \tilde\A^{(2)}_i\X_i;\;...\;\tilde\A^{(k)}_i\X_i]$, where $\tilde\A_i^{(k)}$ specifies whether two nodes in $\G_i$ are \emph{exactly} $k$-hops away; or (3) weighted Layer-wise summation $\Z_i = \alpha_k\sum_{k}\tilde\A^{(k)}_i\X_i$, where $\alpha_k$'s are non-negative weighting that decays with $k$.


Next we consider embedding the substructure instances (i.e., rows of $\Z_i$'s) into a latent space so that statistical manipulations  can better align with the prediction task. The embedding should preserve important proximity relations to facilitate subsequent landmarking: if two substructures are similar, or they often inter-connect with each other, their embedding should be close. In other words, the embedding should be smooth regard to both structural similarities and geometrical interactions. 

A parametric transform on $\Z_i$'s with controlled complexity can guarantee the smoothness of embedding w.r.t. structural similarity, e.g., an autoencoder 
$
f(\Z_i) = \sigma\left(\sigma\left(\Z_i \T_1 + \b_1\right)\T_2+\b_2\right)
$. 
Let $\mathbf{H}_l=f(\Z_i)\in\R^{n_l\times d}$ be the embedding of the $n_l$ sub-graph instances extracted from $\G_l$.
To maintain the smoothness of $\H_i$'s w.r.t. geometric interaction, we will maximize the log-likelihood of the co-ouuurrence of substructure instances in each graph, similar to word2vec \citep{wordvec}
\begin{eqnarray}
\max_{} \sum_{l = 1}^n \sum_{i=1}^{n_l}\sum_{j\in \mathcal{N}^{l}_i} \log\left(\frac{\exp\langle \H_l(i,:),\H_{l}(j,:)\rangle}{\sum_{j'}\exp\langle \H_l(i,:),\H_l(j',:)\rangle}\right) \label{eq:los}
\end{eqnarray}
Here $\H_l(i,:)$ is the $l$th row of $\H_l$, $\langle,\rangle$ is inner product, and $\mathcal{N}^l_i$ are the neighbors of node-$i$ in graph $\G_l$. This loss function tends to embed strongly inter-connecting substructures close to each other. 

\subsection{Substructure Landmarking}

The goal of structural landmarking is to identify a set of informative structural landmarks in the continuous embedding space which has: (1) high statistical coverage, namely, the landmarks should faithfully recover distribution of the substructures from the input graphs, so that we can generalize to new substructure examples from the distribution; and (2) high discriminative power, namely the landmarks should be able to reflect discriminative interaction patterns for classification.

Let $\mathbf{U} = \{ \muu_i, \muu_2,...,\muu_K\}$ be the structural landmarks. In order for them to be representative of the substructure distribution, it is desirable that each sub-graph instance is faithfully approximated with the closest landmark. We will minimize the following distortion loss
\begin{eqnarray}\label{eq:ldmk_loss1}
\sum_{i = 1}^n\sum_{j = 1}^{n_i}\min_{k=1,2,...,K}\|\H_{i}(j,:)-\muu_k\|^2. \end{eqnarray}
Here  $\H_{i}(j,:)$ denotes the $j$th row (substructure) from graph $\G_i$. In practice, we will implement a soft assignment by using one cluster indicator matrix $\W_i\in\R^{n_i\times k}$ for each graph $\G_i$, whose $jk$-th entry is the probability that the $j$th substructure of $\G_i$ belongs to the $k$th landmark $\muu_k$. Inspired by deep embedding clustering \citep{DEC}, $\W_i$ is parameterized by a Student's t-distribution
\begin{eqnarray*}
\W_i(j,k) = \frac{\|(1+\H_i(j,:)-\muu_k\|^2/\alpha)^{-\frac{\alpha+1}{2}}}{\sum_{k'}\|(1+\H_i(j,:)-\muu_k'\|^2/\alpha)^{-\frac{\alpha+1}{2}}}, 
\end{eqnarray*}
and the loss function can be greatly simplified by minimizing the KL-divergence 
\begin{eqnarray}\label{eq:ldmk_loss2}
\min_{\mathbf{U,\H_i's}} \sum_i \text{KL}\left(\W_i,\tilde{\W}_i\right),\;\;\;\; \text{s.t.}\;\; \tilde{\W}_i(j,k) = \frac{\W_i^2(j,k)/\sum_l \W_i(l,k)}{\sum_{k'} \left[\W_i^2(j,k')/\sum_l \W_i(l,k')\right]}.
\end{eqnarray}Here, $\tilde\W_i$ is a self-sharprening version of $\W_i$, and minimizing the KL-distance forces each substructure instance to be assigned to only a small number of landmarks similar to sparse dictionary learning. 
Besides the unsupervised regularization in (\ref{eq:ldmk_loss1}) or (\ref{eq:ldmk_loss2}), learning of the structural landmarks will also be affected by the classification loss, guaranteeing the discriminative power of the landmarks.

\subsection{Identity-Preserving Graph Pooling}

The goal of identity-preserving graph pooling is to project structural details of each graph onto the common space of landmarks, so that a compatible, graph-level feature can be obtained that  simultaneously preserves the identity of the parts (substructures) and models their interactions. 

The structural landmarking mechanism  allows computing rich graph-level features.  
First, we can model substructure distributions. 
The density of the $K$ substructure landmarks in graph $\G_i$ can be computed as  $\p_i = \W_i'\cdot \textbf{1}_{n_i\times 1}$. 
Furthermore, the first-order moment of substructures belonging to each of the $K$ landmarks in  $\G_i$ is $\mathbf{M}_i = \X_i'\cdot\W_i\cdot \P_i^{-1}$ where $\P_i = \text{diag}(\p_i)$, and the $k$th column of $\M_i$ is the mean of $\G_i$'s substructure instances  belonging to the $k$th landmark. 
Second, we can model how the $K$ landmarks interact with each other in graph $\G_i$. 
To do this, we can project the adjacency matrices $\A_i$'s onto the landmark sets and obtain a $\R^{K\times K}$ interaction matrix  
$\mathbf{C}_i = \W_i\cdot\A_i\cdot\W_i'$, which encodes the interacting relations (geometric connections) among the $K$ structural landmarks.

These features can be combined together for final classification. For example, they can be reshaped and concatenated to feed into the fully-connected layer. One can also resort to more intuitive ways; for example, using first-order and second-order features together, one can transform each graph $\G_i$ into a constant-sized, ``landmark'' graph with node feature $\M_i$, node weight $\p_i$, and edge weights $\mathbf{C}_i$. Then standard graph convolution can be applied on the landmark graphs to generate graph-level features (without pains of graph alignment anymore). In experiments, for simplicity, we will compute the normalized interaction matrix
$\tilde{\mathbf{C}}_i = \P_i^{-1}\mathbf{C}_i \P_i^{-1}$ and use it as features, which works pretty well on all the benchmark datasets. More  detailed discussion can be found in Appendix (Sec~8.4 \& 8.7). 


\section{Theoretic Analysis and Discussions}\label{sec:theory}
We provide learning theoretic support on the choice of structural resolution (landmark size $K$). Graphs are bags of inter-connected substructure instances, and each instance $\z$ can be represented by the landmarks as $\mathbf{z} = \sum_{k = 1}^K \alpha_k \muu_k$. A too small number of landmarks fails to recover basic data structures, whereas too many landmarks will result in overfitting (e.g. in exact substructure matching where a maximal $K$ is used for reconstruction) \cite{adpt_size}. 
In dictionary learning, the mutual coherence is a crucial index in evaluating the redundancy of the code-vectors, which is defined as
\begin{eqnarray}\label{eq:ERC}
\mu(\U) = \max_{i,j} \left|\langle \muu_i,\muu_j\rangle\right|,
\end{eqnarray}
where $\langle \cdot,\cdot\rangle$ denotes the normalized correlation.  A lower self-coherence permits
better support recovery \citep{ERC}; 
while large coherence leads to worse stability in both sparse coding and classification \cite{supervised_dic}. In particular, a faithful recovery of the sparse signal support is guaranteed only when 
\begin{eqnarray}
|\alpha|_0 \leq \frac{1}{2}\left(1+\frac{1}{\mu(\U)}\right).\label{eq:cond}
\end{eqnarray}
Obviously, large $\mu(\U)$ leads to unstable solutions. In the following, we quantify a lower-bound of the coherence as a factor of the landmark size $K$ in clustering-based basis selection, since the sparse coding and $k$-means algorithm generate very similar code vectors \citep{clusteringdic}. 
\begin{theorem}
The lower bound of the squared mutual coherence of the landmark vectors increases monotonically with $K$, the number of landmarks in clustering-based sparse dictionary learning. 
\begin{eqnarray*}
\mu^2(\U) 
&\geq& 1 - \frac{4C_dC_p}{u^2_{max}K^{\frac{1}{d}}} \left(\left\lfloor\left(\frac{K}{2}\right)^{\frac{1}{d}} \right\rfloor^{-1}+1\right)
\end{eqnarray*}
Here, $d$ is the dimension, $C_d =  \frac{3}{2}\left(1 + {\log(d)}/{d}\right)\gamma_dV_d$, where $\gamma_d = 1 + d\log(d\log(d))$ and $V_d = 2\Gamma(\frac{1}{2})^d/d\Gamma(\frac{d}{2})$ is the volume of  the $d$-dimensional unit ball; $u_{max}$ is the maximum $\ell_2$-norm of (a subset) of the landmark vectors $\muu_k$'s, and $C_p$ is a factor depending on data distribution $p(\cdot)$. 
\end{theorem}
Proof is in Appendix (Sec~8.1). Theorem~1 says that when the landmark set size $K$ increases, the mutual coherence has a lower bound that consistently increases and violates the recovery condition (\ref{eq:cond}). In fact, a very high structural resolution (like exact matching) leaves a heavy burden to subsequent classifiers by failing to compensate for  structural similarities.  This justifies the SLIM network where the landmark set size can be controlled conveniently to avoid unstable dictionary learning.

\textbf{Discussions}. 
GNNs have shown great potential in graph isomorphism test by generating injective graph embedding, thanks to the theoretic foundations   \cite{GNNpower,WLneural}. However, accurate graph classification needs more thought: classification is not injective; besides,  quality of features is also of notable importance. SLIM provides new insight in both respects: (1) it finds a tradeoff in the duality of handling similarity and distinctness; (2) it explores new ways of generating graph-level features: instead of aggregating all parts together as in GNNs, it taps into the vision of complex systems so that interaction between the parts is leveraged to explain the complexity and improve the learning. More discussions are in Appendix (Sec~8.2-8.8), including 
the choice of spatial/structural resolutions, interpretability, hierarchical and semi-supervised version, and comparison with graph kernels \citep{graph_kernels}.

\section{Experiments}
\label{sec:exp}

\textbf{Benchmark data}. We have used a number of popular benchmark data sets for graph classification. (1) MUTAG: chemical compound data set with 188 instances and two classes; there are 7 node/atom types, and 3 edge/bound types (bond types are ignored). (2) PROTEINS: protein molecule data set with 1113 instances and three classes; there are 3 node types (secondary structure elements). NCI1: chemical compounds data set for cancer cell lines with  4110 instances and two classes. (4) PTC: chemical compound data set for toxicology prediction with 417 instances and 8 classes. (5) D\&D data set for enzyme classification  with 1178 instances and two classes.

\textbf{Competing methods}. We have incorporated a number of highly competitive methods proposed in recent years for comparison: (1) Graph neural tangent kernel (GNTK) \cite{GNTK}; (2) Graph Isomorphism Network (GIN) \cite{GNNpower}; (3) End-to-end graph classification (DCGNN) \cite{DGCNN}; (4) Hierarchical and differential pooling (DiffPool)  \cite{dif_pool}; (5) Self-attention Pooling (SAG) \cite{att_pool}; (6) Convolutional network for graphs (PATCHY-SAN) \cite{10}; 
(7) Graphlet kernel (GK) \cite{GK}; (8) Weisfeiler-Lehman Graph Kernels (WLGK) \cite{WL_kernel}; 9) 
Propagation kernel (PK) \cite{PK}. For method (4),(6),(7),(8),(9) we directly cited their reported results (averaged 10-fold corss-validated error) due to unavailability of their codes; for other competing methods we run their codes with default setting and report the performance.

\textbf{Experimental setting}. We follow the experimental setting in \cite{GNNpower} and \cite{10} and perform 10-fold cross-validation;  we report the average and standard deviation of
validation accuracies across the 10 folds within the cross-validation. In the SLIM network, the spatial resolution is controlled by a BFS with 3-hop neighbors, and the structural resolution is simply set to  $K = 100$; the FC-layer has one hidden layer with dimension 64; cross-entropy id used for classification; weights for the loss term (\ref{eq:los}) and (\ref{eq:ldmk_loss2}) are set to 0.01. No drop-out or batch-normalization is used  considering the size of the benchmark data. 
The hyper-parameters for different dataset include (1) the number of hidden units in the Autoencoder with one hidden unit with a dimension $\{d,d/2, 2d\}$; (2) the optimizer is chosen among SGD or Adagrad, with the learning rate $\{1e-2, 5e-2, 1e-3,5e-3, 1e-4\}$; (3) local graph representation, including node distribution $\A^{(k)}\X_i$, layer-wise distribution, and weighted layer-wise summation (see Sec~3.2 for details); (4) the number of epochs, i.e., a single epoch
with the best cross-validated accuracy averaged over all the 10 folds was selected.  
Overall, a minimal SLIM network is used in the experiments in order to test its performance.

   \begin{wrapfigure}[10]{r}{7cm}
     \centering
     \includegraphics[height=1.15in,width=2.4in]{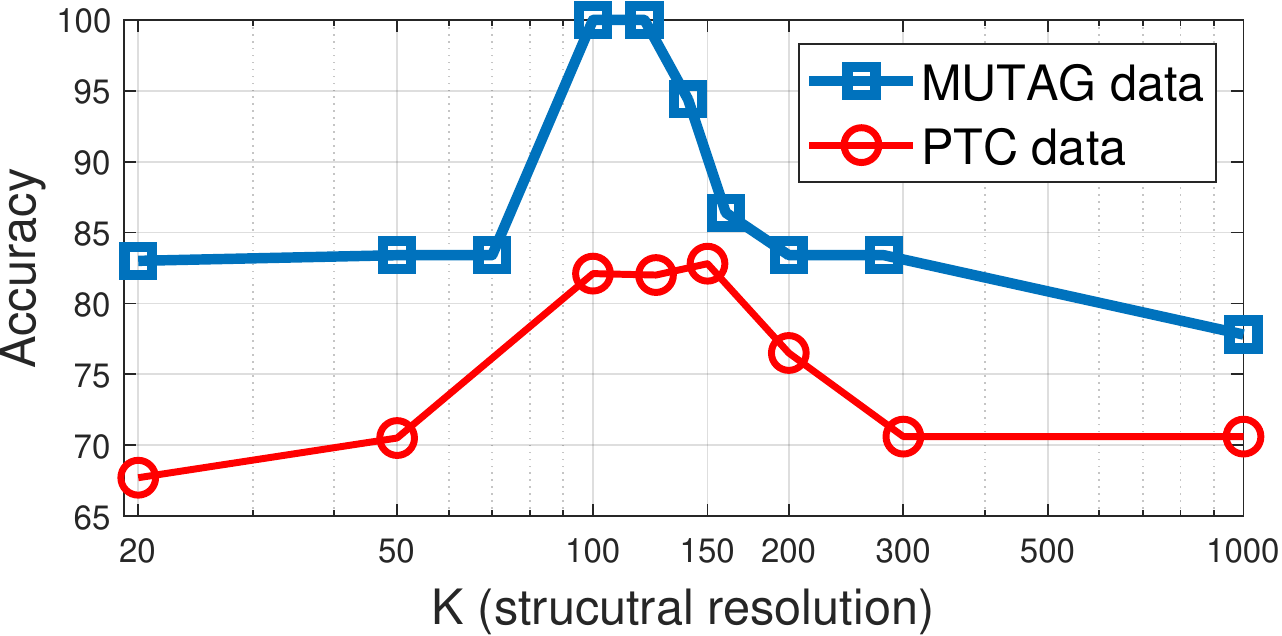}   \vskip -1mm
     \caption{Accuracy vs structural resolution $K$.}\label{fig:dummy}
   \end{wrapfigure}
\textbf{Structural Resolution}. In Figure~\ref{fig:dummy}, we examine the  performance of SLIM under different choices of the structural resolution (landmark set size $K$). As can be seen, the accuracy-vs-$K$ curve has a bell-shaped structure. When $K$ is either too small (underfitting) or too large (coherent landmarks that overfit), the accuracy is low, and the best performance is typically around  a median $K$ value. This validates the correctness of Theorem~1, and  the usefulness of structural landmarking in improving graph classification.



\begin{figure}[ht]
\centering

\subfloat[NCI data.]{ 
             \includegraphics[height=1.5in,width=2.1in]{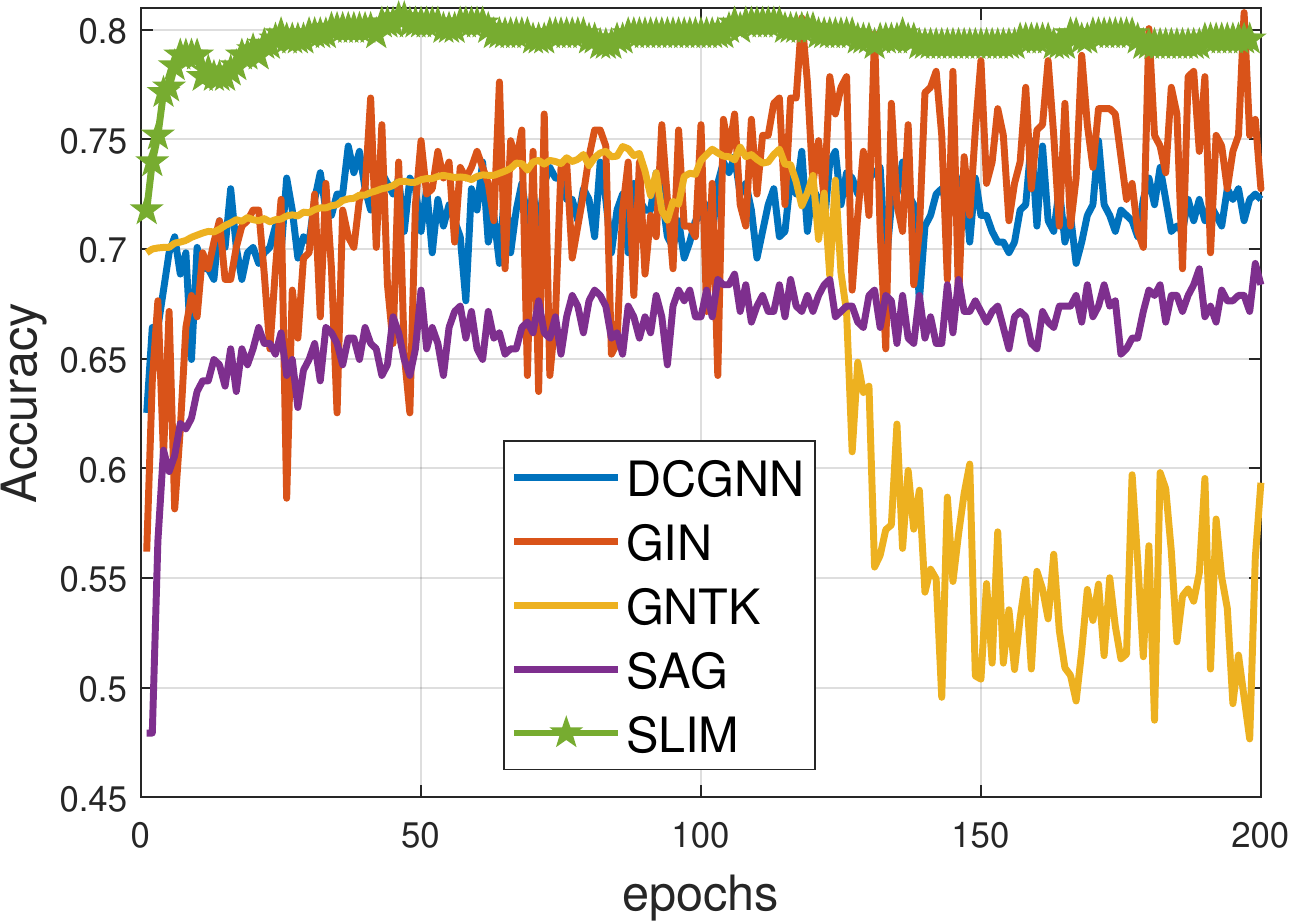}               
 }\hskip 10mm
\subfloat[MUTAG data.]{                               \includegraphics[height=1.5in,width=2.1in]{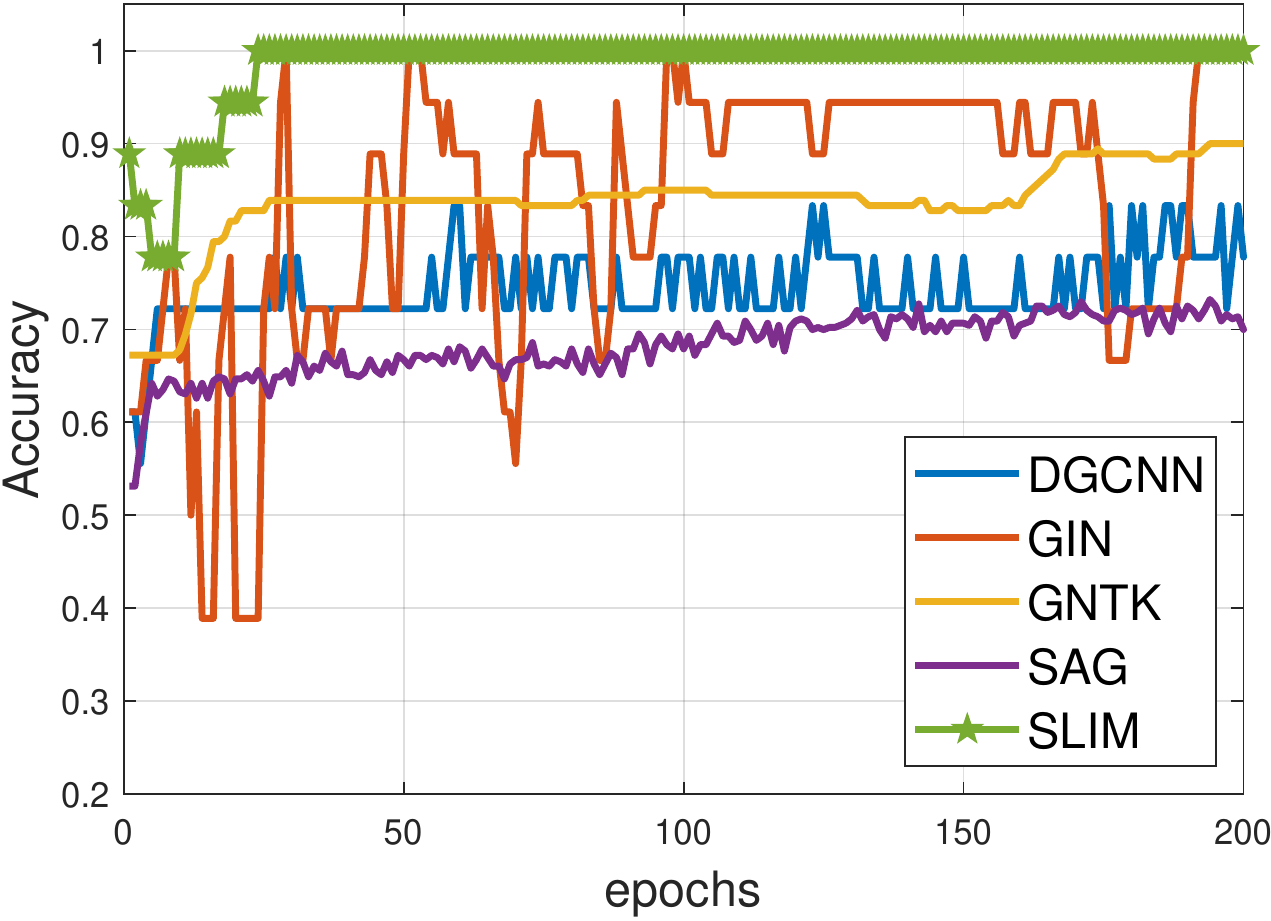}             
}\\

\subfloat[Protein data.]{                                 \includegraphics[height=1.5in,width=2.1in]{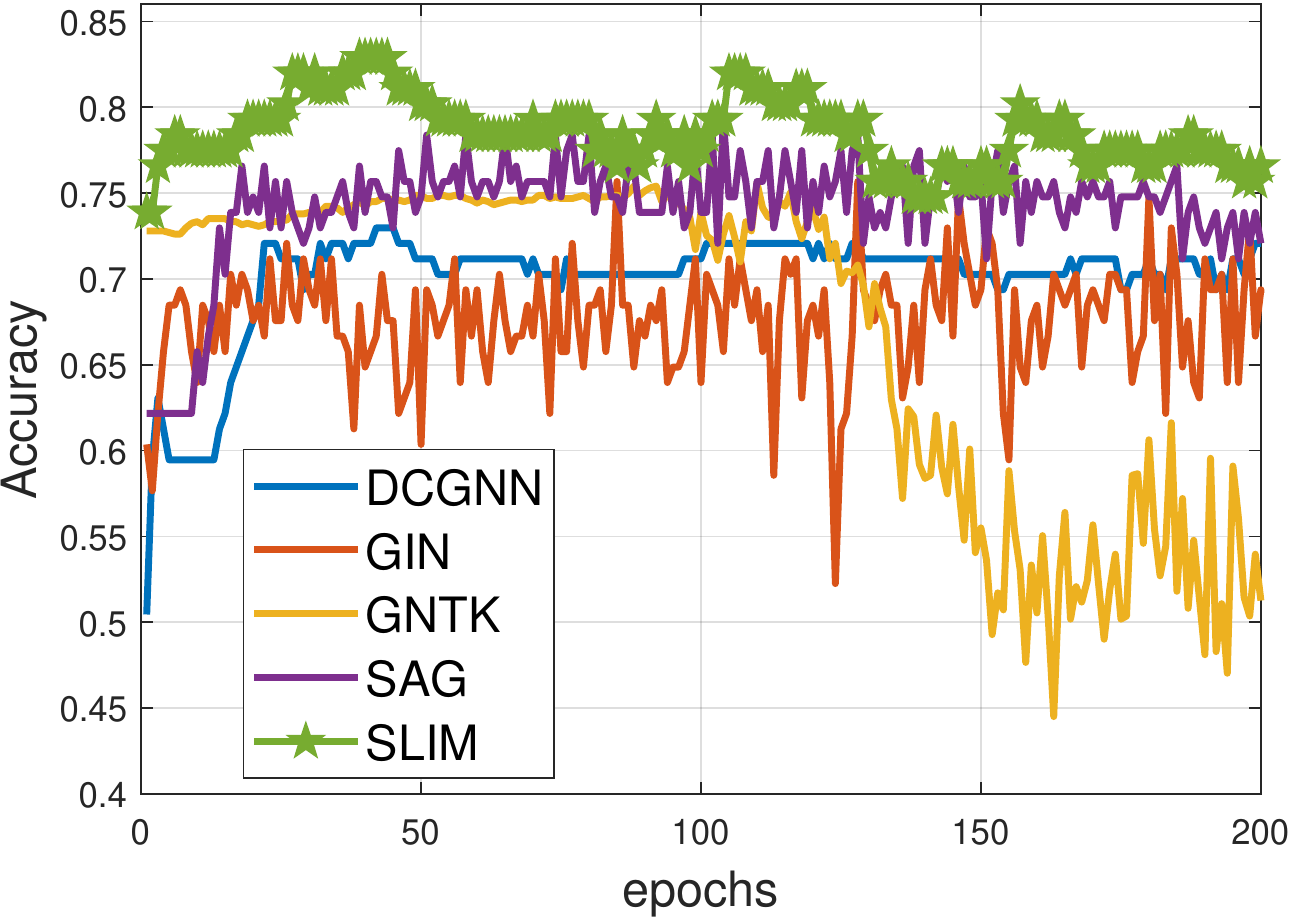}           
}\hskip 10mm
\subfloat[D\&D data.]{                                \includegraphics[height=1.5in,width=2.1in]{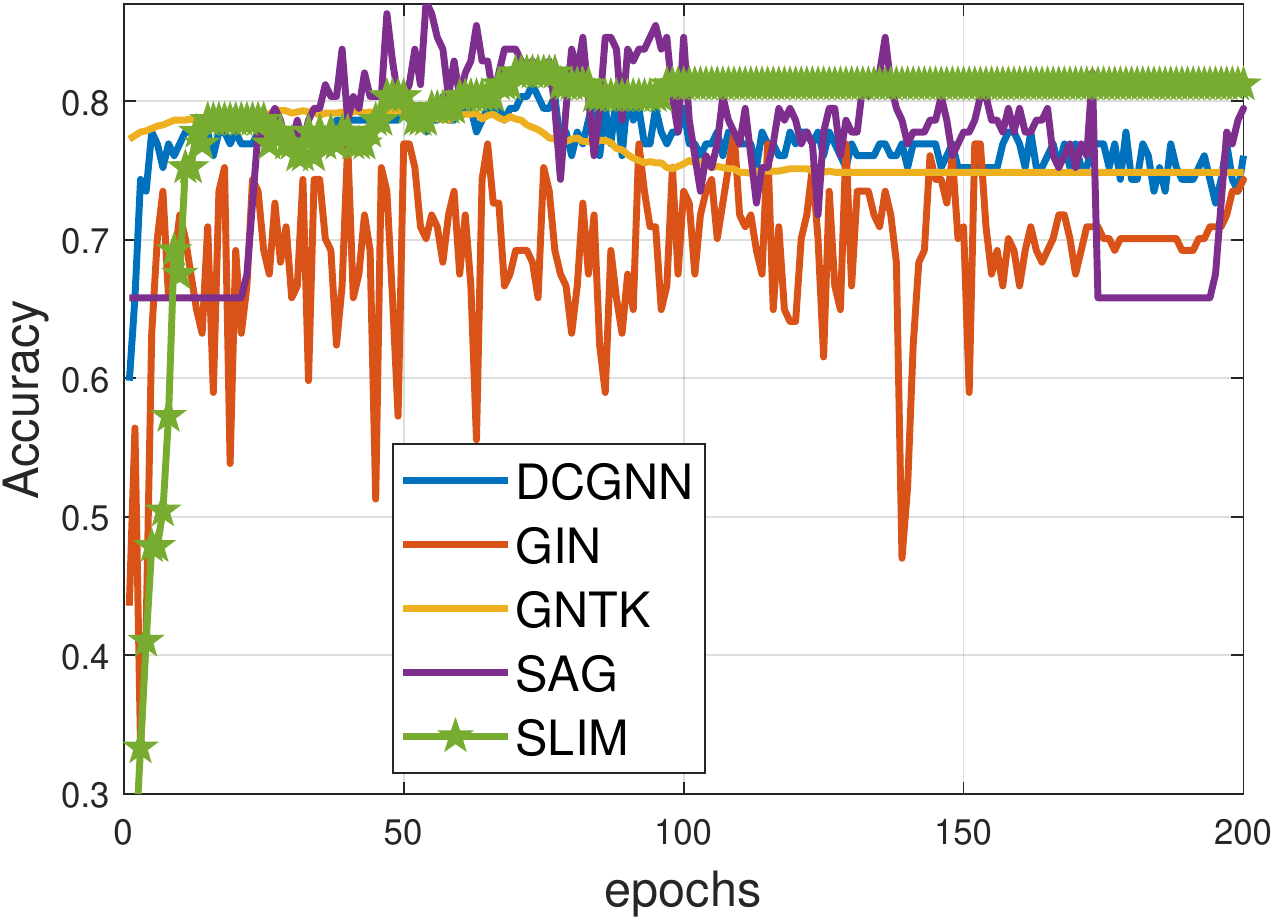}          
}
\caption{Testing accuracy of different algorithms over the training epochs. }
\label{fig:acc}
\end{figure}

\textbf{Classification Performance}. We then compare the performance of different methods in Table~\ref{tbl:acc}. As can be seen, overall, neural network based approaches are more competitive than graph kernels, except that graph kernels have lower fluctuations, and the WL-graph kernel perform the best on the NCI1 dataset.  On most benchmamrk datasets, the SLIM network generates classification accuracies that are either higher or at least as good as other GNN/graph-pooling schemes.

\begin{table}\small
  \caption{Averaged prediction accuracy for different algorithms on 5 benchmark data-sets.}
  \label{sample-table}
  \centering
  \begin{tabular}{lllllll}
    \toprule                
    Category & Algorithm     & MUTAG     & PTC & NCI1 & Protein & D\&D \\
    \midrule
    &GK    &81.38$\pm$1.74 & 55.65$\pm$0.46 & 62.49$\pm$0.27 & 71.39$\pm$0.31 & 74.38$\pm$0.69    \\
   Graph  &PK     & 76.00$\pm$2.69 &  59.50$\pm$2.44 & 82.54$\pm$0.47 & 73.68$\pm$0.68 & 78.25$\pm$0.51\\
   kernel & WLGK & 84.11$\pm$1.91 & 57.97$\pm$2.49& \textbf{84.46$\pm$0.45} & 74.68$\pm$0.49 & 78.34$\pm$0.62\\
        \hline\hline
        &PATCHY-SAN& 92.63$\pm$4.21 & 60.00$\pm$4.82 & 78.59$\pm$1.89& 75.89$\pm$2.76 & 77.12$\pm$2.41\\
    &DGCNN     & 85.83$\pm$1.66 & 68.59$\pm$6.47& 74.46$\pm$0.47 & 75.54$\pm$0.94 & 79.37$\pm$1.03 \\
     &DiffPool     & 90.52$\pm$3.98 & -& 76.53$\pm$2.23 & 75.82$\pm$3.56 & 78.95$\pm$2.40\\
    &GNTK    & 90.12$\pm$8.58 & 67.92$\pm$6.98 & 75.20$\pm$1.53 &75.61$\pm$4.24 & \textbf{79.42$\pm$2.18}\\
GNN    &SAG     & 73.53$\pm$9.68 &  75.67$\pm$3.12 & 74.18$\pm$1.29& 71.86$\pm$0.97 & 76.91$\pm$2.12\\
    &GIN    & 90.03$\pm$8.82 & 76.25$\pm$2.83 & 79.84$\pm$4.57 & 71.28$\pm$2.65 & 77.58$\pm$2.94\\
    & SLIM & \textbf{93.28$\pm$3.36} & \textbf{80.41$\pm$6.92}&	80.53$\pm$2.01& \textbf{77.47$\pm$4.34} & \textbf{79.48$\pm$2.66}\\
    \bottomrule
  \end{tabular}
  \label{tbl:acc}
\end{table}


\textbf{Accuracy Evolution}. We also plot the evolution of the testing accuracy for different methods on the benchmark datasets, so as to have a more comprehensive evaluation on their performance. 
As can be seen from Figure~\ref{fig:acc}, our approach not only generates accurate classification on the benchmark datasets, but also the accuracy  converges relatively faster and remains more stable with respect to the training epochs, making it easier to determine when to stop the training process. Other GNN algorithms can also attain a high accuracy on some of the  benchmark datasets, but the prediction performance fluctuates significantly across the training epochs (even by using large mini-batch sizes). We speculate that stability of the SLIM network arises from explicit modelling of the sub-structure distributions. 
It's also worthwhile to note that on MUTAG data the proposed method produces a classification with 100\% accuracy on more than half of the runs across different folds (Figure~\ref{fig:acc}(b)). It  demonstrates the power of the SLIM network in capturing important graph-level features.

\section{Conclusion} Graph neural networks represent state-of-the-art computational architecture for graph mining.
In this paper, we designed the SLIM network that employs structural landmarking to resolve resolution dilemmas in graph classification and capture inherent interactions in graph-structured systems.  We hope this attempt could open up possibilities in designing  GNNs with informative structural priors.

\bibliography{slim}

\begin{thebibliography}{}

\bibitem[\protect\citename{Alon, }2007]{motif2}
Alon, Uri. 2007.
\newblock Network motifs: theory and experimental approaches.
\newblock {\em Nature Reviews Genetics}, {\bf 8}, 450--461.

\bibitem[\protect\citename{Austin R.~Benson, }2016]{motif4}
Austin R.~Benson, David F.~Gleich, Jure~Leskovec. 2016.
\newblock Higher-order organization of complex networks.
\newblock {\em Science}, {\bf 353}, 163 -- 166.

\bibitem[\protect\citename{Belkin {\em et~al.}, }2006]{mani_reg}
Belkin, Mikhail, Niyogi, Partha, \& Sindhwani, Vikas. 2006.
\newblock Manifold Regularization: A Geometric Framework for Learning from
  Labeled and Unlabeled Examples.
\newblock {\em Journal of Machine Learning Research}, {\bf 7}, 2399--2434.

\bibitem[\protect\citename{Bronstein {\em et~al.}, }2017]{GCN2}
Bronstein, Michael~M., Bruna, Joan, LeCun, Yann, Szlam, Arthur, \&
  Vandergheynst, Pierre. 2017.
\newblock Geometric Deep Learning: Going beyond Euclidean data.
\newblock {\em IEEE Signal Processing Magazine},  18--42.

\bibitem[\protect\citename{Camacho {\em et~al.}, }2018]{biology}
Camacho, D.M., Collins, K.M., Powers, R.K., Costello, J.C., \& Collins, J.J.
  2018.
\newblock Next-Generation Machine Learning for Biological Networks.
\newblock {\em Cell}, {\bf 173}(7), 1581--1592.

\bibitem[\protect\citename{Cangea {\em et~al.}, }2018]{max_pooling}
Cangea, C., Velickovi\'c, P., Jovanovi\'c, N., P., T.~Kipf, \& Li\'o. 2018.
\newblock Towards sparse hierarchical grap classifiers.
\newblock {\em In:} {\em preprint arXiv:1811.01287}.

\bibitem[\protect\citename{Cilliers, }1998]{book_complex2}
Cilliers, Paul. 1998.
\newblock {\em Complexity and Postmodernism: Understanding Complex Systems}.
\newblock Psychology Press.

\bibitem[\protect\citename{Coates \& Ng, }1993]{clusteringdic}
Coates, Adam, \& Ng, Andrew~Y. 1993.
\newblock Learning Feature Representations with K-Means.
\newblock {\em Neural Networks: Tricks of the Trade},  561 -- 580.

\bibitem[\protect\citename{Debarsy {\em et~al.}, }2017]{book_complex}
Debarsy, Nicolas, Cordier, St\'ephane, Ertur, Cem, Nemo, Fran\k{c}ois,
  Nourrit-Lucas, D\'eborah, Poisson, G\'erard, \& Vrain, Christel. 2017.
\newblock {\em Understanding Interactions in Complex Systems, Toward a Science
  of Interaction}.
\newblock Cambridge Scholars Publishing.

\bibitem[\protect\citename{Defferrard {\em et~al.}, }2016]{GCN5}
Defferrard, Micha\"{e}l, Bresson, Xavier, \& Vandergheynst, Pierre. 2016.
\newblock Convolutional Neural Networks on Graphs with Fast Localized Spectral
  Filtering.
\newblock {\em Pages  3844--3852 of:} {\em Advances in Neural Information
  Processing Systems 29}.
\newblock Curran Associates, Inc.

\bibitem[\protect\citename{Donoho \& Huo., }2001]{ERC}
Donoho, D.~L., \& Huo., X. 2001.
\newblock Uncertainty principles and ideal atomic decomposition.
\newblock {\em IEEE Trans. Inf. Theory}, {\bf 47}, 2845--2862.

\bibitem[\protect\citename{Du {\em et~al.}, }2019]{GNTK}
Du, Simon~S, Hou, Kangcheng, Salakhutdinov, Russ~R, Poczos, Barnabas, Wang,
  Ruosong, \& Xu, Keyulu. 2019.
\newblock Graph Neural Tangent Kernel: Fusing Graph Neural Networks with Graph
  Kernels.
\newblock {\em Pages  5723--5733 of:} {\em Advances in Neural Information
  Processing Systems 32}.

\bibitem[\protect\citename{Gao \& Ji, }2019]{unet}
Gao, H., \& Ji, S. 2019.
\newblock Graph u-net.
\newblock {\em In:} {\em Proceedings of the 36th International Conference on
  Machine Learning}.

\bibitem[\protect\citename{Hamilton {\em et~al.}, }2017]{graphSAGE}
Hamilton, William~L., Ying, Rex, \& Leskovec, Jure. 2017.
\newblock Inductive representation learning on large graphs.
\newblock {\em In:} {\em Proceedings of the 31st International Conference on
  Neural Information Processing Systems}.

\bibitem[\protect\citename{Hartwell {\em et~al.}, }1999]{molecular}
Hartwell, L.H., Hopfield, J.J., Leibler, S., \& Murray, A.W. 1999.
\newblock From molecular to modular cell biology.
\newblock {\em Nature}, {\bf 2}(Dec), 47--52.

\bibitem[\protect\citename{Kriegeskorte, }2015]{brain}
Kriegeskorte, Nikolaus. 2015.
\newblock Deep Neural Networks: A New Framework for Modeling Biological Vision
  and Brain Information Processing.
\newblock {\em Annual Review of Vision Science}, {\bf 1}, 417--446.

\bibitem[\protect\citename{Lee {\em et~al.}, }2019a]{motif_embed}
Lee, J.B., Rossi, R.A., Kong, Xiangnan, Kong, X., S.~Kim, E.~Koh, \& Rao, A.
  2019a.
\newblock Graph Convolutional Networks with Motif-based Attention.
\newblock {\em Pages  499--508 of:} {\em Proceedings of the 28th ACM
  International Conference on Information and Knowledge Management}.

\bibitem[\protect\citename{Lee {\em et~al.}, }2019b]{att_pool}
Lee, Junhyun, Lee, Inyeop, \& Kang, Jaewoo. 2019b.
\newblock Self-Attention Graph Pooling.
\newblock {\em Pages  3734--3743 of:} {\em International Conference of Machine
  Learning}.

\bibitem[\protect\citename{Marsousi {\em et~al.}, }2014]{adpt_size}
Marsousi, Mahdi, Abhari, Kaveh, Babyn, Paul~S., \& Alirezaie, Javad. 2014.
\newblock An Adaptive Approach to Learn Overcomplete Dictionaries With
  Efficient Numbers of Elements.
\newblock {\em IEEE Transactions on Signal Processing}, {\bf 62}(12), 3272 --
  3283.

\bibitem[\protect\citename{Mehta \& Gray, }2013]{supervised_dic}
Mehta, Nishant~A, \& Gray, Alexander~G. 2013.
\newblock Sparsity-based generalization bounds for predictive sparse coding.
\newblock {\em Pages  36 -- 44 of:} {\em Proceedings of the 30th International
  Conference on International Conference on Machine Learning}.

\bibitem[\protect\citename{Mikolov {\em et~al.}, }2013]{wordvec}
Mikolov, Tomas, Sutskever, Ilya, Chen, Kai, Corrado, Greg, \& Dean, Jeffrey.
  2013.
\newblock Distributed Representations of Words and Phrases and their
  Compositionality.
\newblock {\em In:} {\em Neural and Information Processing System}.

\bibitem[\protect\citename{Milo1 {\em et~al.}, }2002]{motif1}
Milo1, R., Shen-Orr, S., Itzkovitz, S., Kashtan, N., Chklovskii, D., \& Alon,
  U. 2002.
\newblock Network Motifs: Simple Building Blocks of Complex Networks.
\newblock {\em Pages  824--827 of:} {\em Science}.

\bibitem[\protect\citename{Morris {\em et~al.}, }2019]{WLneural}
Morris, Christopher, Ritzert, Martin, Fey, Matthias, Hamilton, William~L.,
  Lenssen, Jan~Eric, Rattan, Gaurav, \& Grohe, Martin. 2019.
\newblock Weisfeiler and Leman Go Neural: Higher-Order Graph Neural Networks.
\newblock {\em In:} {\em The Thirty-third AAAI Conference on Artificial
  Intelligence}.

\bibitem[\protect\citename{Neumann {\em et~al.}, }2016]{PK}
Neumann, M., Garnett, R., Bauckhage, C., \& Kersting, K. 2016.
\newblock Propagation kernels: efficient graph kernels from propagated
  information.
\newblock {\em Machine Learning}, {\bf 102}(2), 209--245.

\bibitem[\protect\citename{Niepert {\em et~al.}, }2016]{10}
Niepert, Mathias, Ahmed, Mohamed, \& Kutzkov, Konstantin. 2016.
\newblock Learning convolutional neural networks for graphs.
\newblock {\em In:} {\em In Proceedings of The 33rd International Conference on
  Machine Learning}.

\bibitem[\protect\citename{Ron~Meir, }1999]{distortion}
Ron~Meir, Vitaly~Maiorov. 1999.
\newblock Distortion bounds for vector quantizers with finite codebook size.
\newblock {\em IEEE Transactions on Information Theory}, {\bf 45}(5), 1621 --
  1648.

\bibitem[\protect\citename{Schmidt {\em et~al.}, }2019]{material}
Schmidt, Jonathan, Marques, M\'ario R.~G., Botti, Silvana, \& Marques, Miguel
  A.~L. 2019.
\newblock Recent advances and applications of machine learning in solid-state
  materials science.
\newblock {\em NPJ Computational Material}, {\bf 5}, 1581--1592.

\bibitem[\protect\citename{Shervashidze {\em et~al.}, }2009a]{fast-gkernel}
Shervashidze, Nino, Vishwanathan, SVN, Petri, Tobias, Mehlhorn, Kurt, \&
  Borgwardt, Karsten. 2009a (16--18 Apr).
\newblock Efficient graphlet kernels for large graph comparison.
\newblock {\em Pages  488--495 of:} van Dyk, David, \& Welling, Max (eds), {\em
  Proceedings of the Twelth International Conference on Artificial Intelligence
  and Statistics}.
\newblock Proceedings of Machine Learning Research, vol. 5.

\bibitem[\protect\citename{Shervashidze {\em et~al.}, }2009b]{GK}
Shervashidze, Nino, Vishwanathan, SVN, Petri, Tobias, Mehlhorn, Kurt, \&
  Borgwardt, Karsten. 2009b.
\newblock Efficient graphlet kernels for large graph comparison.
\newblock {\em Pages  488--495 of:} {\em Proceedings of the Twelth
  International Conference on Artificial Intelligence and Statistics},  vol. 5.

\bibitem[\protect\citename{Shervashidze {\em et~al.}, }2011]{WL_kernel}
Shervashidze, Nino, Schweitzer, Pascal, van Leeuwen, Erik~Jan, Mehlhorn, Kurt,
  \& Borgwardt, Karsten~M. 2011.
\newblock Weisfeiler-Lehman Graph Kernels.
\newblock {\em Journal of Machine Learning Research}, {\bf 12}(77), 2539--2561.

\bibitem[\protect\citename{Stokes {\em et~al.}, }2020]{drug}
Stokes, Jonathan~M., Kevin~Yang, Kyle~Swanson, Jin, Wengong, Cubillos-Ruiz,
  Andres, Donghia, Nina~M., MacNair, Craig~R., French, Shawn, Carfrae,
  Lindsey~A., Bloom-Ackermann, Zohar, Tran, Victoria~M., Chiappino-Pepe, Anush,
  Badran, Ahmed~H., Andrews, Ian~W., Chory, Emma~J., Church, George~M., Brown,
  Eric~D., Jaakkola, Tommi~S., Barzilay, Regina, \& Collins, James~J. 2020.
\newblock A Deep Learning Approach to Antibiotic Discovery.
\newblock {\em Cell}, {\bf 180}(4), 688--702.

\bibitem[\protect\citename{Velickovic {\em et~al.}, }2017]{gat}
Velickovic, Petar, Cucurull, Guillem, Casanova, Arantxa, Romero, Adriana, Lio,
  Pietro, \& Bengio, Yoshua. 2017.
\newblock Graph Attention Networks.
\newblock {\em In:} {\em International Conference on Learning Representations}.

\bibitem[\protect\citename{Vishwanathan {\em et~al.}, }2010]{graph_kernels}
Vishwanathan, S.V.N., Schraudolph, Nicol~N., Kondor, Risi, \& Borgwardt,
  Karsten~M. 2010.
\newblock Graph Kernels.
\newblock {\em Journal of Machine Learning Research}, {\bf 11}(40), 1201--1242.

\bibitem[\protect\citename{Wang {\em et~al.}, }2019]{SEED}
Wang, Lichen, Zong, Bo, Ma, Qianqian, Cheng, Wei, Ni, Jingchao, Yu, Wenchao,
  Liu, Yanchi, Song, Dongjin, Chen, Haifeng, \& Fu, Yun. 2019.
\newblock Inductive and Unsupervised Representation Learning on Graph
  Structured Objects.
\newblock {\em In:} {\em International COnference on Learning Representations}.

\bibitem[\protect\citename{Wernicke, }2006]{motif3}
Wernicke, S. 2006.
\newblock Efficient Detection of Network Motifs.
\newblock {\em IEEE/ACM Transactions on Computational Biology and
  Bioinformatics}, {\bf 3}(4), 347--359.

\bibitem[\protect\citename{Wu {\em et~al.}, }2020]{GNNreview}
Wu, Zonghan, Pan, Shirui, Chen, Fengwen, Long, Guodong, Zhang, Chengqi, \& Yu,
  Philip~S. 2020.
\newblock A Comprehensive Survey on Graph Neural Networks.
\newblock {\em IEEE Transactions on Neural Networks and Learning Systems}.

\bibitem[\protect\citename{Xie {\em et~al.}, }2016]{DEC}
Xie, J., Girshick, R., \& Farhadi, A. 2016.
\newblock Unsupervised deep embedding for clustering analysis.
\newblock {\em In:} {\em International Conference on Learning Representations}.

\bibitem[\protect\citename{Xu {\em et~al.}, }2019]{GNNpower}
Xu, Keyulu, Hu, Weihua, Leskovec, Jure, \& Jegelka, Stefanie. 2019.
\newblock How Powerful are Graph Neural Networks?
\newblock {\em In:} {\em International Conference on Learning Representations}.

\bibitem[\protect\citename{Yanardag \& Vishwanathan, }2015]{Deep-gkernel}
Yanardag, Pinar, \& Vishwanathan, SVN V~N. 2015.
\newblock Deep Graph Kernels.
\newblock {\em Page  1365–1374 of:} {\em Proceedings of the 21th ACM SIGKDD
  International Conference on Knowledge Discovery and Data Mining}.

\bibitem[\protect\citename{Yang {\em et~al.}, }2018]{GNNmotif1}
Yang, Carl, Liu, Mengxiong, Zheng, Vincent~W., \& Han, Jiawei. 2018.
\newblock Node, Motif and Subgraph: Leveraging Network Functional Blocks
  Through Structural Convolution.
\newblock {\em In:} {\em IEEE/ACM International Conference on Advances in
  Social Networks Analysis and Mining}.

\bibitem[\protect\citename{Ying {\em et~al.}, }2018]{dif_pool}
Ying, Rex, You, Jiaxuan, Morris, Christopher, Ren, Xiang, Hamilton, William~L.,
  \& Leskovec, Jure. 2018.
\newblock Hierarchical Graph Representation Learning with Differentiable
  Pooling.
\newblock {\em Page  4805–4815 of:} {\em Proceedings of the 32nd
  International Conference on Neural Information Processing Systems}.

\bibitem[\protect\citename{Zaheer {\em et~al.}, }2017]{deep_set}
Zaheer, Manzil, Kottur, Satwik, Ravanbakhsh, Siamak, Poczos, Barnabas,
  Salakhutdinov, Russ~R, \& Smola, Alexander~J. 2017.
\newblock Deep Sets.
\newblock {\em Pages  3391--3401 of:} {\em Advances in Neural Information
  Processing Systems 30}.
\newblock Curran Associates, Inc.

\bibitem[\protect\citename{Zhang {\em et~al.}, }2018]{DGCNN}
Zhang, Muhan, Cui, Zhicheng, Neumann, Marion, \& Chen, Yixin. 2018.
\newblock An End-to-End Deep Learning Architecture for Graph Classification.
\newblock {\em In:} {\em The Thirty-Second AAAI Conference on Artificial
  Intelligence}.

\bibitem[\protect\citename{Zhang {\em et~al.}, }2020]{GNNreview3}
Zhang, Z., Cui, P., \& Zhu, W. 2020.
\newblock Deep Learning on Graphs: A Survey.
\newblock {\em IEEE Transactions on Knowledge and Data Engineering}.

\bibitem[\protect\citename{Zhou {\em et~al.}, }2018]{GNNreview2}
Zhou, Jie, Cui, Ganqu, Zhang, Zhengyan, Yang, Cheng, Liu, Zhiyuan, \& Sun,
  Maosong. 2018.
\newblock Graph Neural Networks: A Review of Methods and Applications.
\newblock {\em In:} {\em ArXiv}.
\newblock Arixv.

\end{thebibliography}
\bibliographystyle{authordate1}
\newpage
\section{Appendix}


\subsection{Proof of Theorem I}
\begin{proof}
Suppose we have $n$ spatial instances embedded in the $d$-dimensional latent space as $\{\z_1, \z_2,...,\z_n\}$, and the landmarks (or codevectors) are defined as $\muu_1,\muu_2,...,\muu_K$. Let $p(\z)$ be the density function of the instances. Define the averaged distance between the instance and the closest landmark point as
\begin{eqnarray}
s = \frac{1}{n}\sum_{i = 1}^n \|\z_i - \muu_{c(i)}\|_2,
\end{eqnarray}
where $c(i)$ is the index of the closest landmark to instance $i$. As expected, $s$ will decay with the number of landmarks with the following rate \citep{distortion}
\begin{eqnarray}
s \leq C_d C_p\left(\left\lfloor\left(\frac{K}{2}\right)^{\frac{1}{d}} \right\rfloor^{-1}+1\right)K^{-\frac{1}{d}}
\end{eqnarray}
where $C_d$ is a dimension-dependent factor $C_d =  \frac{3}{2}\left(1 + \frac{\log(d)}{d}\right)\gamma_d$, with $V_d = 2\Gamma(\frac{1}{2})^d/d\Gamma(\frac{d}{2})$ the volume of  the unit ball in k-dimensional Euclidean space and $\gamma_d = 1 + d\log(d\log(d))$; $C_p = \left(\int p(\z)^{\frac{d}{d+1}}d\z\right)^{\frac{d+1}{d}}$ is a factor depending on the distribution $p$. 

Since $s$ is the average distortion error, we can make sure that there exists a non-empty subset of instances ${\Omega_z}$ such that $\|\z_i - \muu_{c(i)}\|\leq s$ for $i\in \Omega_z$. Next we will only consider this subset of instances and the relevant set of landmarks will be denoted by $\Omega_u$. For the landmarks $\muu_p\in \Omega_u$, we make a realistic assumption that there are enough instances so that we can always find one instance $\z$ falling in the middle of $\muu_p$ and its closest landmark neighbor $\muu_p$. In this case, we have then bound the distance between the closest landmark pairs as 
\begin{eqnarray*}
\|\muu_p - \muu_q\|\leq \|\muu_p - \z\|_2 + \|\muu_q - \z\|_2\leq 2 s. 
\end{eqnarray*}

For any such pair, assume that the angle spanned by them is $\theta_{pq}$. we can bound the angle between the two landmark vectors by
\begin{eqnarray}
\sin\left(\theta_{pq}\right)\leq \frac{2s}{\|\muu_p\|}.
\end{eqnarray}
Let $u_{max} = \max_{\muu_p\in\Omega_u} \|\muu_p\|_2$, we can finally low-bound the normalized correlation between close landmark pairs, and henceforth the coherence of the landmarks, as 
\begin{eqnarray*}
\mu^2(\U) &\geq& \displaystyle \max_{p,q\in \Omega_u} \cos^2(\theta_{pq})\\
&=& \max_{p,q\in \Omega_u} {1 - \sin^2(\theta_{pq})^2}\\
&\geq & 1 - \frac{4s^2}{u^2_{max}}\\
&\geq& 1 - \frac{4C_dK^{-\frac{1}{d}}}{u^2_{max}} \left(\left\lfloor\left(\frac{K}{2}\right)^{\frac{1}{d}} \right\rfloor^{-1}+1\right)
\end{eqnarray*}
This indicates that the squared mutual coherence of the landmarks has a lower bound that consistently increases when the number of the landmark vectors, $K$, increases in a dictionary learning process. 
\end{proof}
This theorem provides important guidance on the choice of  structural resolution. It shows that when a clustering-based dictionary learning scheme is used to determine the structural landmarks, the size of the dictionary $K$ can not be chosen too large; or else the risk of overfitting can be huge. Note that exact sub-structure matching as is often practiced in current graph mining tasks corresponds to an extreme case where the number of landmarks, $K$, equals the number of unique sub-structures; therefore it should be avoided in practice. The structural landmarking scheme is a flexible framework to tune the number of landmarks, and to avoid overfitting.  


\subsection{Choice of Spatial and Structural Resolutions}

The spatial resolution determines the ``size'' of the local sub-structure (or sub-graph), such as functional modules in a molecule. Small sub-structures can be very limited in terms of their representation power, while too large sub-structures can mask the right scale of the local components crucial to the learning task. An optimal spatial resolution can be data-dependent. In practice, we will restrict the size of the local sub-graphs to 3-hop BFS neighbors, considering that the ``radius'' of the graphs in the benchmark data-sets are usually around 5-8.  We then further fine-tune the spatial resolution by assigning a  non-negative weighting on the nodes residing on different layers from the central code in the local subgraph. Such weighting is shared across all the sub-graphs and can be used to adjust the importance of each layer of the BFS-based sub-graph. The weighting can be chosen as a monotonously decaying function, or optimized through learning. 

The choice of structural resolution has a similar flavor in that too small or too large resolutions are neither desirable. On the other hand, it can be adjusted conveniently by tuning the landmark set size $K$ based on the validation data. In our experiments, $K$ can be chosen by cross validation; for simplicity, we fix $K = 100$.

Finally, note that geometrically larger substructures (or sub-graphs) are characterized by higher variations among instances due to the exponential amount of configuration. Therefore, the structural resolution should also commensurate with spatial resolutions. For example, substructures constructed by 1-hop-BFS may use a smaller landmark size $K$ than those with 3-hop-BFS. In our experiments we do not consider such dependencies yet, but will study it in our future research.  

\subsection{Comparison with Graph Kernels}
Graph kernels are powerful methods to measure the similarity between graphs.
The key idea is to compare the sub-structure pairs from the two graphs and compute the accumulated similarity, where examples of substructures include random walks, paths, sub-graphs, or sub-trees. Among them, paths/sub-graphs/sub-trees are  deterministic sub-structures in a graph, while random walks are stochastic sequences (of nodes) in a graph.  

Although the SLIM network considers sub-structures as the basic processing unit, it has a number of important differences compared with graph kernels. First, we consider optimizable sub-structural landmarks, which is dependent on the class labels and therefore discriminative; in comparison, the sub-structures considered in graph kernels are identified by enumerating or sampling among a large amount of pre-determined candidates. Second, the similarity measured by graph kernels is between a apir of sub-structures across the two graphs; in comparison, the SLIM network models the interacting relation in each graph as its feature.
Third, it can be difficult to interpret graph kernels due to the nonlinearity of kernel methods and the exponential amount of sub-structures; in comparison, the SLIM network maintains a reasonable amount of ``landmark'' structures and so can provide informative clues on the prediction result.

\subsection{Hierarchical Version}
\subsubsection{Subtlety in Spatial Resolution Definition}
First we would like to clarify a subtlety in the definition of spatial resolutions. In physics, resolution is defined as the smallest distance (or interval) between two objects that can be separated; therefore it must involve two scales: the scale of the object, and the scale of the interval.  
Usually these two scales are proportional. In other words, you cannot have a large intervals and small objects, or the opposite (a small interval and large object). For example, in the context of imaging, each object is a pixel and the size of the pixel is the same as the interval between two adjacent pixels.

In the context of graphs, each object is a sub-graph centered around one node, whose scale is manually determined by the order of the BFS-search centered around that node. Therefore, the interval between two sub-graphs may be smaller than the size of the sub-graph. For example, two nodes $i$ and $j$ are direct neighbors, and each of them haa a 3-hop sub-graph. Then, the interval between these two subgraphs, if defined by the distance between $i$ and $j$, will be 1-hop; this is smaller than the size of the two sub-graphs, which is 3-hop. In other words, the two objects/subgraphs indeed overlap with each other, and the scale of the object and the scale of the interval between objects is no longer commensurate (large objects and small interval in this scenario).

This scenario makes it less complete to define spatial resolutions just based on the size of the sub-graphs (as in the main text), since there are actually two scales to define. To avoid unnecessary confusions, we skip these details. In practice, one has two choices dealing with the discrepancy: (1)  requiring that the sub-graphs are not overlapping, i.e., we do not have to grow one $k$-hop subgraph around each node; instead, we just explore a subset of the sub-graphs. This can be implemented in a hierarchical version which we discuss in the next subsection; (2) we still allow each node to have a local sub-graph and study them together, which helps cover the diversity of subgraphs since theoretically, an ideal choice of the subgraph is highly domain specific and having more sub-graph examples gives a better chance to include those sub-graphs that are beneficial to the prediction task. 

\subsubsection{Hierarchical SLIM}

We can implement a hierarchical version of SLIM so that sub-graphs of different scales, together with the interacting relation between sub-graphs under each scale, can be captured for final prediction. Note that in \cite{dif_pool} a hierarchical clustering scheme is used to partition one graph, in a bottom up manner, to less and less clusters. We can implement the same idea and construct a hierarchy of scales each of which will host a number of sub-structures. The structural landmarking scheme will be implemented in each layer of the hierarchy to generate graph-level features specific to that scale. Finally these features can be combined together for graph classification. 
   
\subsection{Semi-supervised SLIM Network}
The SLIM network is flexible and can be trained in both fully supervised setting and semi-supervised  setting. This is because the SLIM model takes a parametric form and so it is inductive and can generalize to any new samples; on the other hands, the  clustering-based loss term in (\ref{eq:ldmk_loss2}) can be evaluated on both labeled samples and unlabeled samples, rendering the extra flexibility to look into the distribution of the testing sample in the training phase, if they are available. This is in flavor very similar to the smoothness constraint widely used in semi-supervised learning, such as the graph-regularized manifold learning \cite{mani_reg}.  Therefore, the SLIM network can be implemented in the following modes
\begin{itemize}
\item Supervised version. Only training graphs and their labels are available during the training phase, and the loss function (\ref{eq:ldmk_loss2}) is only computed on the training samples. 

\item Semi-supervised version. Both labeled training graphs and unlabeled testing graphs are available. The loss function (\ref{eq:ldmk_loss2}) will be computed on both the training and testing graphs, wile the classification loss function will only be evaluated on the training graph labels. 
\end{itemize}

\subsection{Interpretability} The SLIM network not only generates accurate prediction in graph classification problems, but can also provide important clues on interpreting the prediction results, because the graph-level features in SLIM bear clear physical meaning. For example, assume that we use the interaction matrix $\mathbf{C}_i$ for the $i$th graph $\G_i$ as its feature representation; and the $pq$th entry then quantifies the connectivity  strength between the $p$th sub-structure landmark and the $q$th structure landmark. Then, by checking the $K^2$-dimensional model coefficients from the fully-connected layer, one can then tell which subset of substructure-connectivity (i.e., two substructures are directly connected in a graph) is important in making the prediction. To improve the interpretability one can further imposes a sparsity constraint on the model coefficient. 

In traditional graph neural networks such as GraphSAGE of GIN, node features are transformed through many layers and finally mingled altogether through graph pooling. The resultant graph-level representation, whose dimension is manually determined and each entry pools the values across all the nodes in the graph,  can be difficult to interpret. 

\subsection{The prediction Layer}

The SLIM network renders various possibilities to generate the prediction layer. 
\begin{itemize}
    \item Fully connected layer. The interaction matrix can be re-shaped into a vector, or transformed to a smaller matrix via bilateral dimension reduction before reshaped into a vector. Then a fully connected layer follows for the final prediction.
    \item Landmark Graph. Each graph $\G_i$ can be transformed into a landmark-graph a with fixed number of $K$ (landmark) nodes, with $\mathbf{p}_i$ and $\mathbf{C}_i$ quantifying the weight of each node and the edge between every pair of nodes, and ${\M}_i$ the feature of each node (see definition in Section3.3). Then, this graph can be subject to a graph convolution such as $\mathbf{D}_i^{-1}\mathbf{C}_i\M_i$ generate a fixed-dimensional graph-level feature without having to take care of the varying graph size. We will study this in our future experiments. 
    
    \item Riemannian manifold. When using the interaction matrix $\mathbf{C}_i$ or the normalized version as graph level features, we can treat each graph as a point in the Riemannian manifold due to the symmetry and positive semi-definiteness of the representation. Then the distance between two interaction matrices can be computed as the Wasserstein distance between two Gaussian distributions with the interaction matrix as covariances, which has a closed-form. We will study this in our future experiments.  
    
\end{itemize}

\subsection{Interaction versus Integration}
The SLIM network and existing GNNs represent two different flavors of learning, namely, interaction modelling versus integration approach. Interaction modelling is based on mature understanding of complex systems and can provide physically meaningful interpretations or support for graph classification; integration based approaches bypass the difficulty of preserving the identity of sub-structures and instead focus on whether the integrated representation is an injective mapping, as typically studied in  graph isomorphism testing. 

Note that an ideal classification is different from isomorphism testing and is  not injective. In a good classifier, the goal of deciding which samples are similar and which are not are equally important.
Here comes the tradeoff between handling similarity and distinctness. The Isomorphism-flavor GNN's are aimed at preserving the differences between local sub-structures (even just a very minute difference), and then map the resultant embedding to the class labels. Our approach, on the other hand, tries to absorb patterns that are sufficiently close to the same landmark, and then map the landmark-based features to class labels. In the latter case, the structural resolution can be tuned in a flexible way to explore different fineness levels, thus tuning the balance between ``similarity'' and ``distinctness''; in the meantime,  the structural landmarks allow preserving sub-structure identities and exploiting their interactions.

\end{document}